\pdfoutput=1

\documentclass[11pt]{article}

\usepackage[review]{coling}

\usepackage{times}
\usepackage{latexsym}

\usepackage[T1]{fontenc}

\usepackage[utf8]{inputenc}

\usepackage{microtype}

\usepackage{inconsolata}

\usepackage{graphicx}
\usepackage{amsthm}
\usepackage{amsmath,amssymb}
\usepackage{subfigure}
\usepackage{multirow}

\usepackage{authblk}

\title{BoRA: Bi-dimensional Weight-Decomposed Low-Rank Adaptation}

\author{\textbf{Qiushi Wang}}
\author{\textbf{Yuchen Fan}}
\author{\textbf{Junwei Bao}}
\author{\textbf{Hongfei Jiang}}
\author{\textbf{Yang Song}}
\affil{Zuoyebang Education Technology Co., Ltd.}

\begin{document}
\maketitle
\begin{abstract}
In recent years, Parameter-Efficient Fine-Tuning (PEFT) methods like Low-Rank Adaptation (LoRA) have significantly enhanced the adaptability of large-scale pre-trained models. Weight-Decomposed Low-Rank Adaptation (DoRA) improves upon LoRA by separating the magnitude and direction components of the weight matrix, leading to superior performance. However, DoRA’s improvements are limited to the vertical dimension, resulting in an asymmetrical pattern between horizontal and vertical dimensions. This paper introduces BoRA, an innovative extension of LoRA and DoRA, characterized by symmetrical properties across horizontal and vertical dimensions. Our approach optimizes the weight matrix symmetrically by adjusting both column-wise and row-wise magnitudes. Extensive experiments demonstrate that BoRA surpasses state-of-the-art PEFT methods, including LoRA and DoRA, achieving superior results across various benchmarks.
\end{abstract}

\section{Introduction}

\begin{figure}[h]
\centering
\includegraphics[scale=0.4]{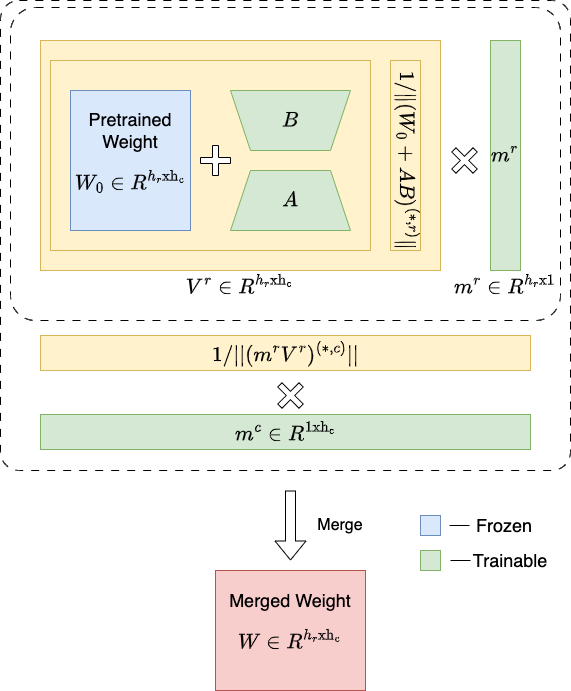}
\caption{Structure of BoRA: blue indicates frozen parameters, green indicates trainable parameters. Low-rank adaptation matrices and two independent magnitude matrices ensure symmetrical adjustment of the weight matrix.}
\label{fig:ess}
\end{figure}

Recent advancements in the field of natural language processing (NLP) have been mainly driven by large language models (LLMs)(\citet{devlin2018bert}, \citet{radford2019language}, \citet{brown2020language}, \citet{raffel2020exploring}). These models built with billions of parameters, have demonstrated remarkable capabilities in understanding and generating human language. However, the deployment and fine-tuning of these models come with significant computational and storage requirements, making them less feasible for widespread use, especially in resource-constrained environments.

Parameter-Efficient Fine-Tuning (PEFT) methods have emerged as promising solutions to address these challenges. By fine-tuning only a small subset of the model parameters while keeping the rest frozen, PEFT methods enable the adaptation of LLMs to specific tasks with minimal computational overhead. Among these techniques, Low-Rank Adaptation (LoRA)\citep{hu2021lora} stands out. LoRA enhances the fine-tuning of large language models by optimizing a low-rank representation of weight updates, reducing computational costs and retaining model performance.

Building on the foundations laid by LoRA, a new method called Weight-Decomposed Low-Rank Adaptation (DoRA)\citep{liu2024dora} has been proposed. By decomposing the weight into magnitude and direction components, DoRA achieves a more efficient adaptation process. This method has shown promising results, offering a markable progress on model performance. However, the improvements of DoRA are constrained to vertical dimension, resulting in a disparity in symmetry compared to other fine-tuning methods.

In this paper, we introduce a novel approach that further pushes the boundaries of Parameter-Efficient Fine-Tuning. We term our method BoRA, drawing inspiration from LoRA and DoRA while integrating novel enhancements to overcome challenges and improve performance. The BoRA method decouples magnitude components from directional variations in the weight matrix through a low-rank adaptation representation and two independent, trainable magnitude parameters, as illustrated in Figure \ref{fig:ess}. These magnitude parameters symmetrically influence column and row dimensions within the weight matrix. By integrating bi-dimensional weight matrix adjustments with small-scale parameter tuning, BoRA demonstrates superior performance across various tasks while preserving computational efficiency.

Our contributions can be summarized as follows:

\begin{itemize}
\item[$\bullet$] We introduce BoRA, an advanced variant of LoRA and DoRA. Through comprehensive experiments, we demonstrate that BoRA consistently outperforms state-of-the-art PEFT methods across multiple benchmark datasets.

\item[$\bullet$] A set of metrics, correlated with the timestep, are utilized to illustrate the bi-dimensional dynamics of the weight matrix, thus providing a comprehensive evaluation of the model's behavior. This methodology offers advanced perspectives for forthcoming analysis of modeling paradigms.

\end{itemize}

\section{Related Works \label{related}}

\subsection{PEFT}

Parameter-Efficient Fine-Tuning (PEFT) represents a strategic approach to adapting pretrained deep learning models for new or customized tasks without training all model parameters. This technique is characterized by its ability to deliver strong performance through small-scale weight adjustments. 

Parameter-Efficient Fine-Tuning (PEFT) can be broadly classified into three categories, Additive methods, Selective methods and Reparametrization-based methods\citep{lialin2023scaling}. Additive Methods include two main approaches, Adapters(\citet{houlsby2019parameter}, \citet{pfeiffer2020adapterhub}, \citet{zhu2021counter}, \citet{pfeiffer2020adapterfusion}) and Soft Prompts(\citet{radford2019language}, \citet{liu2023gpt}, \citet{lester2021power}, \citet{li2021prefix}). Selective Methods involve fine-tuning only specific layers of the network, as demonstrated in \citet{donahue2014decaf}, \citet{gheini2021cross}, \citet{zaken2021bitfit}, \citet{vucetic2022efficient}. 
Reparametrization-Based Methods, such as LoRA (Low Rank Adaptation)\citep{hu2021lora}, leverage low-rank representations to minimize the number of trainable parameters \citep{maddox2020rethinking, li2018measuring, arora2018stronger, malladi2023kernel}. The modular design of LoRA allows for its integration into various frameworks to enhance performance \citep{huang2023lorahub, liu2023moelora}.

\subsection{LoRA}

Low-Rank Adaptation (LoRA)\citep{hu2021lora} facilitates the training of specific trainable parameters while keeping the pre-trained weights frozen. The key idea of LoRA is utilizing low-rank matrices for approximating the matrix-adjusting progression during the full parameter fine-tuning phase. For a linear layer with a weight matrix denoted as $W_0 \in \mathbb{R}^{d1 \times d2}$, LoRA incorporates two additional low-rank matrices denoted as $A \in \mathbb{R}^{d1 \times rank}$ and $B \in \mathbb{R}^{rank \times d2}$. Both these matrices exhibit a hidden dimension of $rank$, markedly smaller than the dimensions of $W_0$. While training, original weight matrices remain fixed, and the only trainable parameters are matrices $A$ and $B$. Ultimately, the matrices $A$ and $B$ are added on the original weight, so the final weight matrices can be represented as $W_0+AB$. This approach significantly reduces the number of parameters that need to be trained during fine-tuning.

\subsection{Weight normalize}

Weight Normalization\citep{salimans2016weight} is designed to improve the training of neural networks by reparameterizing the weight vectors of neural network layers. The main idea behind weight normalization is to separate the magnitude and direction of the weight vectors. By decoupling the magnitude and the direction, weight normalization allows the optimization process to more effectively adjust the direction of the weight vector while separately modulating its scale. This enhancement improves the conditioning of the gradient, thereby advancing the convergence of the optimization process. Furthermore, it accelerates the convergence of the stochastic gradient descent algorithm.

\subsection{DoRA}

Weight-Decomposed Low-Rank Adaptation (DoRA)\citep{liu2024dora} represents a synthesis of the distinctive features of both LoRA and weight normalization techniques. The implementation of a small-scale trainable parameter benefits on minimizing computational complexity. Furthermore, the decoupling of the weight matrix's magnitude and direction components augments the training process and achieves performance comparable to full-parameter fine-tuning without any additional inference latency compared to LoRA. A detailed exploration of DoRA will be presented in subsequent section \ref{DoRA}.

\section{Method}

\subsection{Feature Weighting in Weight Matrix \label{DoRA}}

In the DoRA method, the magnitude dimension is modulated by a trainable parameter matrix $m$, while the direction dimension is modified by the trainable matrices $A$ and $B$. However, the trainable matrix $m$ only influences the columns of the weight matrix. Let $\mathbf{d} \in \{r, c\}$ represents the dimension, which can be either row ($r$) or column ($c$). Consequently, $m \in \mathbb{R}^{1 \times h_c}$. Notably, the pre-trained weight matrix, denoted as $W_0 \in \mathbb{R}^{h_r \times h_c}$, remains fixed during the training process. 

Subsequently, the normalization operation ensures that each column of the matrix $\frac{W_0 + AB}{\lVert (W_0 + AB) ^{(*,c)}\rVert}$ is converted into a unit vector, preserving its direction while standardizing its magnitude. Here, $(*,c)$ denotes each vector along the column dimension, referring to every column in the weight matrix. The expression $\lVert (W_0 + AB) ^{(*,c)}\rVert$ denotes a one-dimensional vector in $\mathbb R^{1 \times h_c}$, where each element of this vector corresponds to the norm value of the respective column in the matrix $W_0+AB$. Thus, we define $V^c$ as  $V^c = \frac{W_0 + AB}{\lVert (W_0 + AB) ^{(*,c)}\rVert}$, where every column vector in $V^c$ maintains unit length. The matrix $m$ adjusts the scale of each column within the normalized weight matrix $V^c$ through a element-wise multiplication operation. 

Thus, the final weight matrix of DoRA, denoted as $W$, can be elaborated as :
\begin{equation}
W = m\frac{W_0 + AB}{\lVert (W_0 + AB) ^{(*,c)}\rVert} = m V^c
\end{equation}

Within the framework of a linear layer, the operation of matrix multiplication is simply formalized by the functional mapping $Y = XW^T + b$. Upon excluding consideration of the bias vector, the resultant matrix may be represented as:
\begin{equation}
Y_{ij} = \sum_{l=1}^{h_c}X_{il}m_{1l}V^c_{jl}
\end{equation}

Upon examining the relationship between the elements within magnitude matrix $m$ and the input matrix $X$, it becomes evident that each corresponding scalar in $m$ is multiplied to individual elements within each specific column of the input matrix $X$. For instance, when $l=1$, each element in the first column of the input matrix $X$ is multiplied by the first scalar $m_{11}$. In NLP tasks, each row in the input matrix corresponds to a token in a sequence, with columns representing the features that define each token's embedding in the vocabulary\citep{pennington2014glove, mikolov2013distributed}. Viewing columns within this matrix as distinct features, it is observed that $m$ plays an important role in weighting these features within the input matrix. This modulation effectively recalibrates the distribution of each input feature, thereby optimizing the result output.

\subsection{Bi-dimensional Weight-Decomposed Low-Rank Adaptation}

However, DoRA merely introduces an additional operator upon the input feature, which may sometimes prove insufficient. Through the application of various metrics, we have observed that the adjustments made by DoRA are asymmetrical, potentially resulting in an uneven training process. These metrics will be elaborated upon in Section \ref{sec: metrics}.

We propose our novel approach called Bi-dimensional Weight-Decomposed Low-Rank Adaptation (BoRA). This method enables the magnitude adjustment of the weight matrix in two distinct dimensions, $\mathbf{d} \in \{r, c\}$, ensuring a comprehensive adjustment of weight matrix.

Our approach is comprised of two procedures. 
\begin{itemize}
\item Initially, let us denote the pre-trained weight matrix as $W_0 \in \mathbb R^{h_r \times h_c}$. The weight matrices $A$ and $B$ are used to adjust the pre-trained weight matrix, resulting in the new weight matrix, $W_0 + AB$. Subsequently, we execute normalization procedures on its horizontal axis. We designate the resultant matrix post-normalization as $V^r$, wherein every row in the matrix $V^r$ constitutes a unit vector. Consequently, the process of normalization may be formalized as:
$$V^r = \frac{W_0+AB}{\rVert (W_0 + AB)^{(*,r)} \rVert}$$
, where $\rVert (W_0 + AB)^{(*,r)} \rVert$ represents a vector in $\mathbb R^{h_r \times 1}$, whose elements are the norm values of the corresponding rows of the weight matrix $W_0 + AB$.
Subsequently, we introduce a matrix $m^r \in \mathbb R^{h_r \times 1}$, conceptualized as a trainable parameter set, endowed with the capacity to calibrate the magnitude of individual rows within the matrix $V^r$. Therefore, the ultimate expression derived from the first step can be represented as  $m^rV^r$.

\item In Step 2, we normalize each column of the matrix $m^rV^r$ such that each becomes a unit vector. Let $H^c$ represent the matrix obtained after this column normalization. The normalization process can be expressed as follows:
$$H^c = \frac{m^rV^r}{\lVert (m^rV^r)^{(*,c)} \rVert}$$
, where $\lVert (m^rV^r)^{(*,c)} \rVert$ represents a vector in $\mathbb R^{1 \times h_c}$, whose elements are the norm values of the corresponding columns of the weight matrix $m^rV^r$.
Ultimately, the adjustable weight matrix, denoted as $m^c \in \mathbb R^{1 \times h_c}$, calibrate the scale of individual columns within the weight matrix $H^c$. The proposed technique can be methodically articulated as follows: $W = m^cH^c$
\end{itemize}

In conclusion, our method can be formalized as follows:

\begin{equation}
W = m^c \frac{\frac{W_0 + AB}{\lVert (W_0 + AB)^{(*,r)} \rVert} m^r}{\lVert (\frac{W_0 + AB}{\lVert (W_0 + AB)^{(*,r)}\rVert} m^r)^{(*, c)}\rVert}
\end{equation}

Therefore, following multiplication by a linear layer, the resulting output matrix is:

\begin{equation}
\begin{aligned}
Y_{ij} &= \sum_{l=1}^{h_c}X_{il}m^c_{1l}H^c_{jl}\\&=\sum_{l=1}^{h_c}X_{il}m^c_{1l}\frac{m^r_{j1}V^r_{jl}}{\sqrt{\sum_{k=1}^{h_r}(m^r_{k1}V^r_{kl})^2}}
\end{aligned}
\end{equation}

When utilizing only the column magnitude matrix $m^c$ as DoRA, each element in matrix is applied to the corresponding column in the input matrix $X$, effectively adjusting the weights of the input features. However, examining each column in the output matrix $Y$, the weighting operation fails to make any impact on different output columns. Although DoRA facilitates deeper training by adjusting the weights of the input features, the consistency of the weighting across each output feature limits its potential. To address this limitation, the BoRA framework introduces an additional row magnitude weight matrix $m^r$, which diversifies the feature weighting operation for each output feature, corresponding to each column of the output matrix. This modification may enhance the training process, leading to improved model performance and deeper network training efficacy. Consequently, the ulitization of both column and row magnitude matrices in BoRA may offer a further enhanced framework for complex model training scenarios, driving advancements in the optimization capabilities of neural networks.

\subsection{Symmetry in Rows and Columns \label{sec: metrics}}

\begin{figure*}[htbp]
\centering
\subfigure[Full Parameter]{
    \rotatebox{90}{\scriptsize{~~~~~~Column-wise}}
    \includegraphics[width=0.2\linewidth]{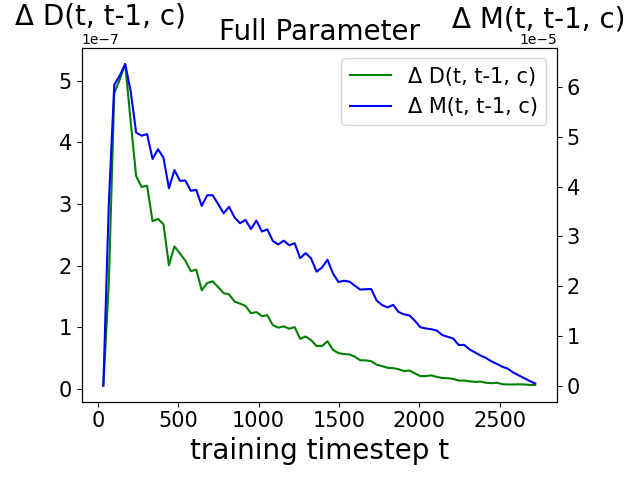}}
\subfigure[LoRA]{
    \includegraphics[width=0.2\linewidth]{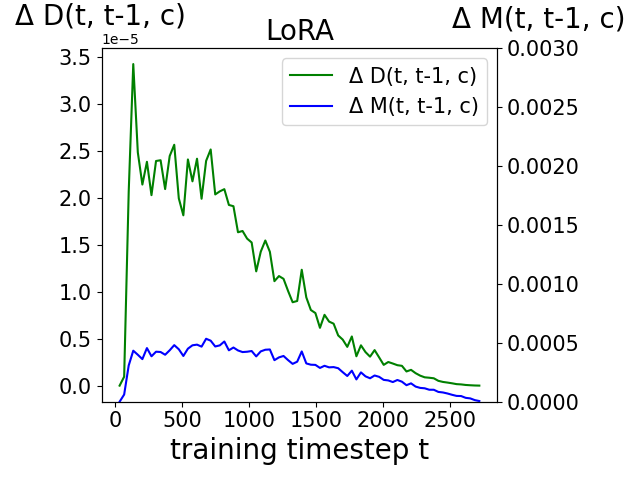}}
\subfigure[DoRA]{
    \includegraphics[width=0.2\linewidth]{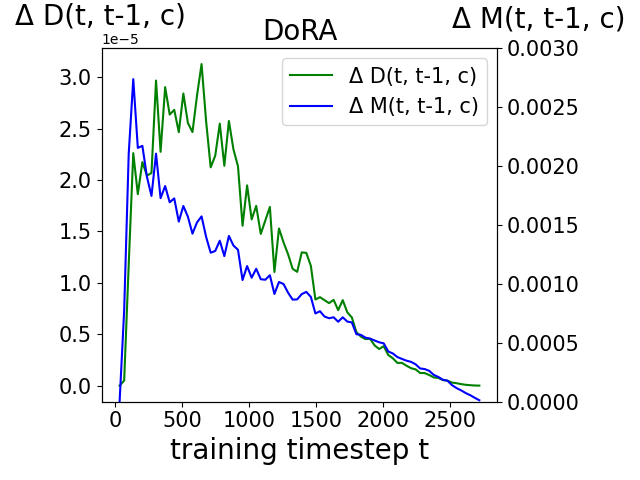}}
\subfigure[BoRA]{
    \includegraphics[width=0.2\linewidth]{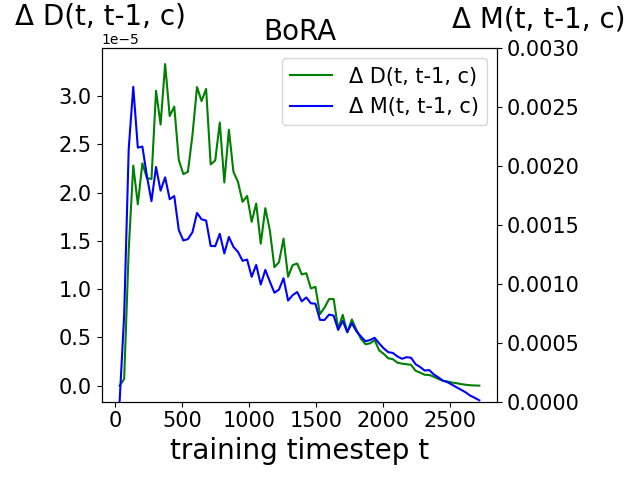}}

\subfigure[Full Parameter]{
    \rotatebox{90}{\scriptsize{~~~~~~~~~~Row-wise}}
    \includegraphics[width=0.2\linewidth]{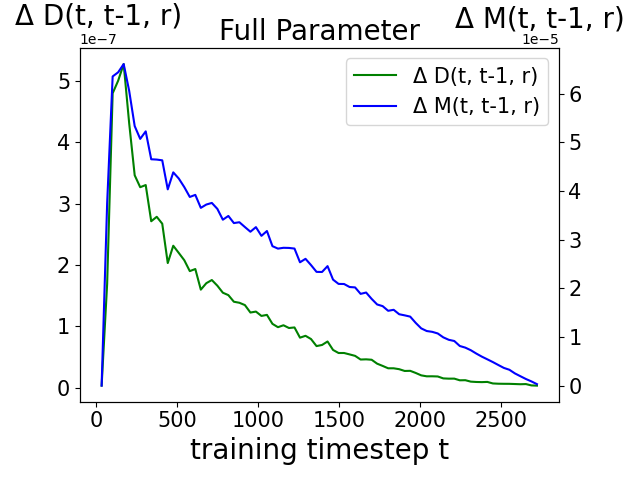}}
\subfigure[LoRA]{
    \includegraphics[width=0.2\linewidth]{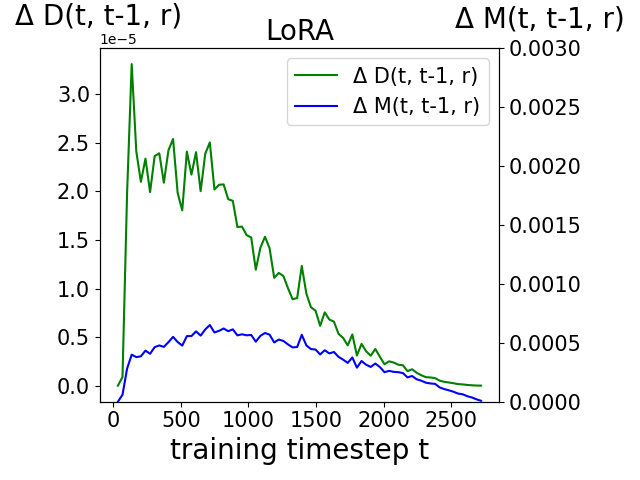}}
\subfigure[DoRA]{
    \includegraphics[width=0.2\linewidth]{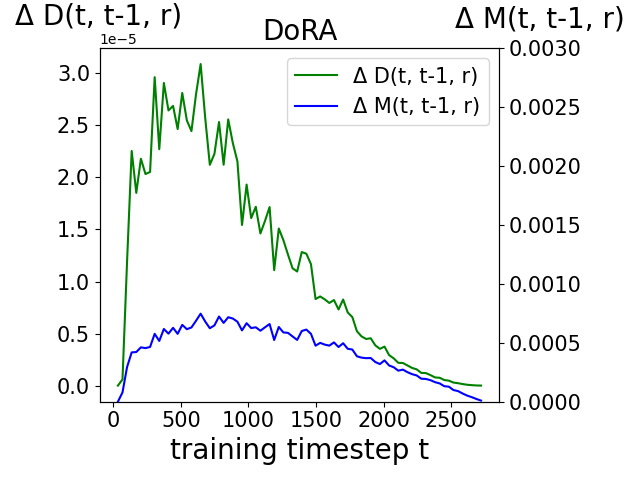}}
\subfigure[BoRA]{
    \includegraphics[width=0.2\linewidth]{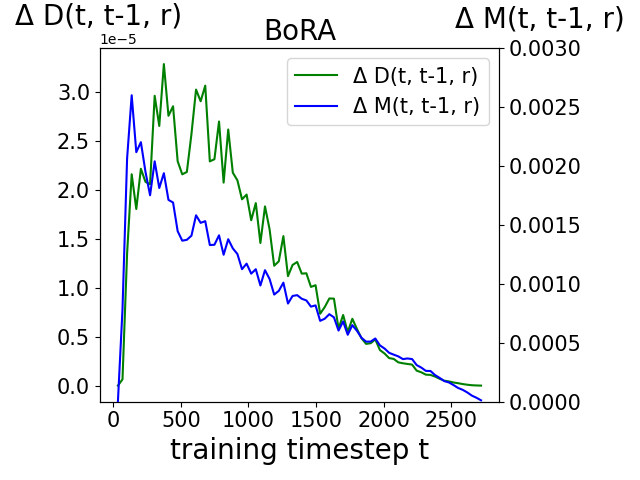}}
    
\caption{Magnitude and direction updates from time interval $t-1$ to $t$, examining full parameter tuning (a, e) and three PEFT methods: LoRA (b, f), DoRA (c, g), and BoRA (d, h). Each method is evaluated in two dimensions: column-wise and row-wise. Green represents direction changes, while blue represents magnitude changes.}
\label{tab:fig1}
\end{figure*}

We employ a multi-metric analysis strategy to evaluate the training dynamics of various model training techniques. Our investigation revealed noteworthy correlations within magnitude/direction changings and timesteps. Thereby shedding new light on the intricate mechanisms underlying the efficacy of model tuning practices.

\begin{figure*}[htbp]
\centering
\subfigure[Full Parameter]{
    \rotatebox{90}{\scriptsize{~~~~~~Column-wise}}
    \includegraphics[width=0.2\linewidth]{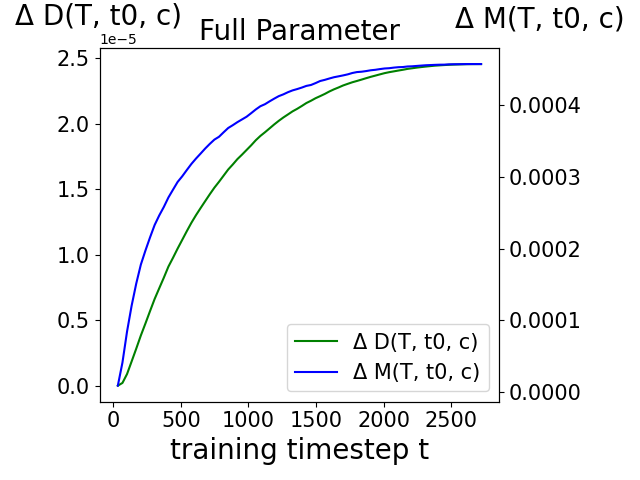}}
\subfigure[LoRA]{
    \includegraphics[width=0.2\linewidth]{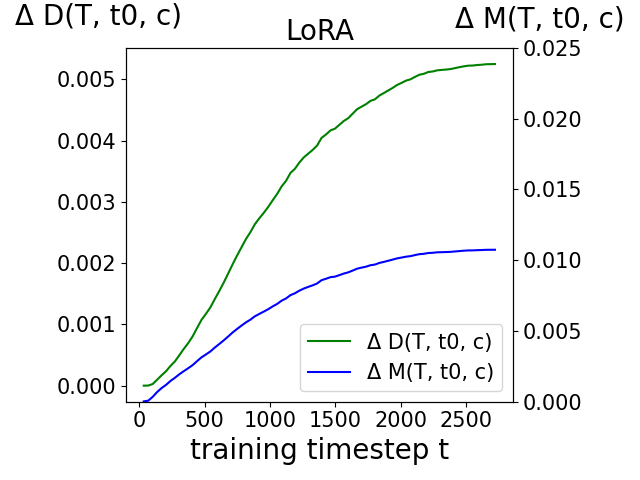}}
\subfigure[DoRA]{
    \includegraphics[width=0.2\linewidth]{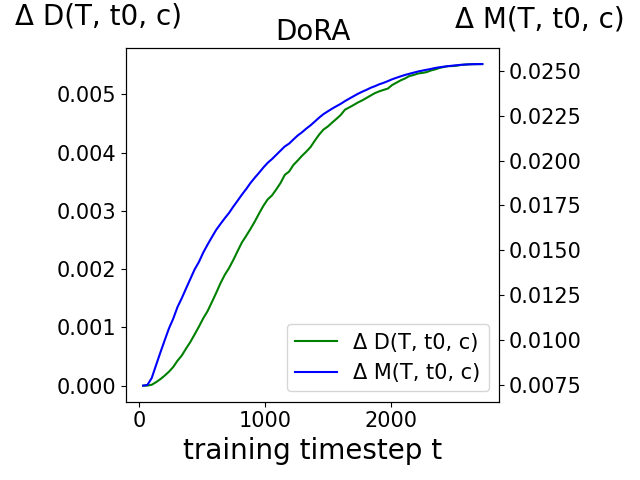}}
\subfigure[BoRA]{
    \includegraphics[width=0.2\linewidth]{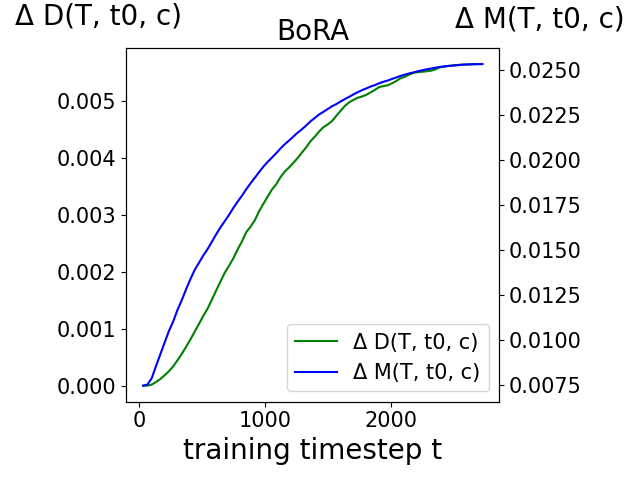}}

\subfigure[Fullp Parameter]{
    \rotatebox{90}{\scriptsize{~~~~~~~~~~Row-wise}}
    \includegraphics[width=0.2\linewidth]{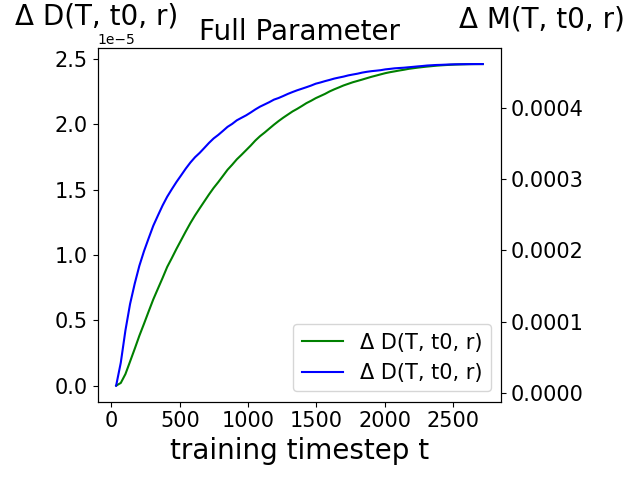}}
\subfigure[LoRA]{
    \includegraphics[width=0.2\linewidth]{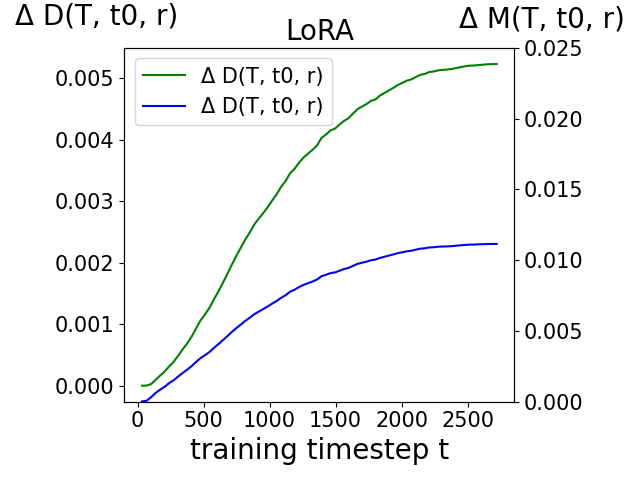}}
\subfigure[DoRA]{
    \includegraphics[width=0.2\linewidth]{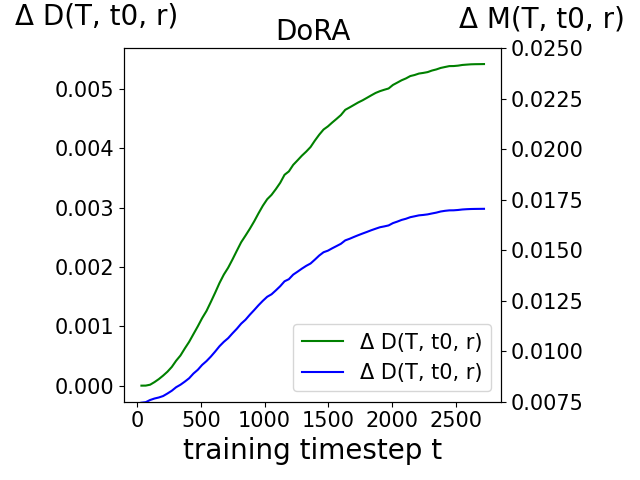}}
\subfigure[BoRA]{
    \includegraphics[width=0.2\linewidth]{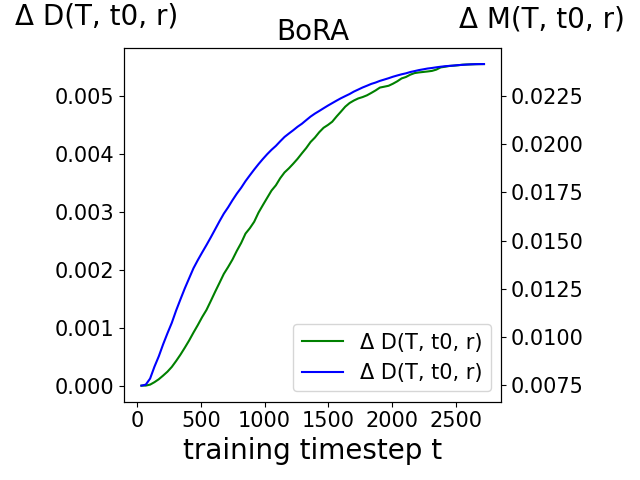}}

\caption{Magnitude and direction updates post-training, examining full parameter tuning (a, e) and three PEFT methods: LoRA (b, f), DoRA (c, g), and BoRA (d, h). Each method is evaluated in two dimensions: column-wise and row-wise. Green represents direction changes, while blue represents magnitude changes.}
\label{tab:fig2}
\end{figure*}

Due to the substantial gap of the size between the trainable magnitude matrix and the weight matrix of the base model, representing the pattern of metrics for LLMs presents significant challenges. To analyze the behavior of the weight matrix, we trained the RoBERTa (base) model\citep{liu2019roberta} on the CoLA dataset\citep{wang2018glue} for 80 epochs using various methods. The results are represented in \ref{limit}. In figures \ref{tab:fig1} and \ref{tab:fig2}, we present the average value of the metrics on the query weight matrix across all 12 layers of the model.

To introduce the temporal dimension into this analysis, we examine the differences in magnitude and direction of merged weight between two distinct timesteps. Let $W_{t1}$ and $W_{t2} \in \mathbb{R}^{h_c \times h_r}$ represent the final merged weight matrices at times $t1$ and $t2$, respectively. The magnitude distance and the direction distance are defined as follows:
\begin{equation}
\triangle M(t1, t2, \mathbf{d}) = \frac{\sum_{i=1}^{h_\mathbf{d}}|\lVert W_{t1}^{(i,\mathbf{d})} \rVert - \lVert W_{t2}^{(i,\mathbf{d})} \rVert|}{h_\mathbf{d}}
\end{equation}
\begin{equation}
\triangle D(t1, t2, \mathbf{d}) = \frac{\sum_{i=1}^{h_\mathbf{d}}(1 - cos(W_{t1}^{(i,\mathbf{d})}, W_{t2}^{(i,\mathbf{d})}))}{h_\mathbf{d}}
\end{equation}
, where $\triangle M$ quantitatively encapsulates the magnitude change in the parameter between the discrete intervals $t1$ and $t2$. Concurrently, $\triangle D$ is conceptualized as the integrated directional change, representing the directional difference in the same temporal framework. 

Furthermore, $W_{t1}^{(i,\mathbf{d})}$ and $W_{t2}^{( i,\mathbf{d})}$ denote the $i$-th vector on $\mathbf{d}$ dimension at timestep $t1$ and $t2$, $\mathbf{d} \in \{r, c\}$. For instance, when analyzing the weight matrix by its columns, which corresponds to setting $\mathbf{d} = c$, the notation $W^{(i,c)}$ specifically refers to the $i$-th column of the weight matrix $W$. Concurrently, $\lVert W_{t1}^{(i,\mathbf{d})} \rVert$ and $\lVert W_{t2}^{(i,\mathbf{d})} \rVert$ illustrate the magnitude value of the $i$-th vector on $\mathbf{d}$ dimension of the corresponding weight matrix. The function $cos(\cdot,\cdot)$ refers to the application of the cosine similarity metric, commonly used to measure angles between vectors in multidimensional spaces. By adopting this method, we aim to present a coherent and reliable analysis of the distance in magnitude and direction between matrices on two different time steps.

We first examine the alterations in magnitude and direction between successive timesteps, specifically comparing timestep $t$ with the preceding timestep $t-1$. The objective is to elucidate the sequential variations in magnitude and direction during the training process. The metrics $\triangle M(t, t-1, \mathbf{d})$ and $\triangle D(t, t-1, \mathbf{d})$ are represented in Figure \ref{tab:fig1}.

As illustrated in Figure \ref{tab:fig1}, an examination across both columns and rows reveals a consistent pattern in the training process when utilizing full parameter training. The variations in the magnitude and direction during the full parameter training process exhibit a similar pattern vertically and horizontally.

A similar trend can be observed in the training process of LoRA, where the metrics demonstrate a consistent pattern across columns and rows. This consistency is also evident in BoRA. Contrarily, the pattern diverges when analyzing columns and rows in the context of DoRA training. 

By constraining both the range of $\triangle M(t, t-1, \mathbf{d})$ and $\triangle D(t, t-1, \mathbf{d})$, we observe that LoRA exhibits reduced variability in magnitude changes across both columns and rows. However, the introduction of a trainable matrix $m$ specific to columns, as implemented in DoRA, results in a significant magnitude change in the column-wise perspective.

In our proposed method, BoRA, we introduce two trainable matrices $m^c$ and $m^r$ for the magnitudes of the two dimensions: columns and rows, thereby aligning the behavioral patterns of both columns and rows. BoRA facilitates a similar modulation of magnitudes in both dimensions, akin to the patterns observed in LoRA and full-parameter training. By introducing matrices $m^c$ and $m^r$ , BoRA also amplifies the magnitude changes in both columns and rows.

We also investigate the distances in both magnitude and direction between the weight matrices of a fine-tuned model and its base model. Here, let $T$ represent the final training step and $t_0$ denote the initial training step. Consequently, we analyze the metrics $\triangle M(T, t_0, \mathbf{d})$ and $\triangle D(T, t_0, \mathbf{d})$, as illustrated in Figure \ref{tab:fig2}, to capture these changes.

Examining Figure \ref{tab:fig2}, we observe a maintained consistency across columns and rows in both $\triangle M(T, t_0, \mathbf{d})$ and $\triangle D(T, t_0, \mathbf{d})$ during full parameter training and upon the application of LoRA. It is noteworthy that the application of LoRA results in a markedly reduced magnitude alternation as opposed to the use of DoRA(column-wise) and BoRA, while the directional variation is comparably preserved. Utilizing DoRA, the isolation of magnitude tuning of columns from direction tuning amplifies the magnitude within columns, while concurrently maintaining low values comparable to LoRA within rows. 

Introducing bi-dimensional trainable magnitude matrices, BoRA schematizes a symmetrical pattern across both columns and rows, thereby enhancing the alterations of magnitudes in both dimensions. Consequently, this approach may lead to enhanced model performance through the refined tuning of magnitudes, maintaining highly coherent and efficient parameter interactions across both dimensions.

\section{Experiments \label{sec: exp}}

\subsection{Setup}

In this section, we conduct a variety of experiments to demonstrate the efficacy of our fine-tuning method. Specifically, we assess the performance of our approach BoRA, in comparison to several state-of-the-art Parameter-Efficient Fine-Tuning (PEFT) methods: LoRA, DoRA, rsLoRA(\citep{kalajdzievski2023rankstabilizationscalingfactor}), AsyLoRA(\citep{zhu2024asymmetrylowrankadaptersfoundation}), PiSSA(\citep{meng2024pissaprincipalsingularvalues}) and MoRA(\citep{jiang2024morahighrankupdatingparameterefficient}).

Our investigation includes an evaluation of BoRA alongside those baseline methods utilizing large language models, including Llama-2-7b, Mistral-7b-v0.1 and Llama-3-8b\citep{touvron2023llama, jiang2023mistral, llama3modelcard}. To demonstrate the efficacy of our method, we conduct a comparative analysis of these models in the context of instruction-tuning.

Initially, we evaluate the models' performance on Natural Language Generation (NLG) tasks to provide a broad overview of their general capabilities. We fine-tuned models using a uniform instructional dataset WizardLM-Evol-Instruct-70k\citep{xu2024wizardlm} and subsequently evaluated on the MT-Bench dataset\citep{zheng2024judging}. Our evaluative approach incorporates the assessments by GPT-4\cite{achiam2023gpt}.

Subsequently, we assess their performance on Natural Language Understanding (NLU) tasks to offer a more comprehensive and detailed evaluation of their overall effectiveness. The Commonsense Reasoning Dataset\citep{hu2023llm} is an extensive corpus encompassing eight distinct tasks, each associated with its own separate test datasets. These varied tasks are strategically designed to probe distinct facets of the models' capabilities, and ultimately achieves the evaluation of their generalized performance. We utilize the Commonsense Reasoning Dataset for training purposes, subsequently evaluating performance on the designated test set for each individual task. 

Ultimately, we select a generic task to conduct the ablation study. Here, the Llama-2-7b model, which was trained on the WizardLM-Evol-Instruct-70k dataset and evaluated using the MT-Bench dataset, serves as our focus. Throughout this process, our aim is to reveal the distinctions and the relative advantages of our proposed fine-tuning method.

\subsection{Performance on NLG tasks}

To evaluate the performance of fine-tuning methods, we fine-tuned the model with different PEFT methods, and quantified the corresponding answers of models to specific instructions in MT-Bench dataset employing of GPT-4, which provided a scoring spectrum from 1 to 10. A comparative analysis was then conducted to elucidate the performance efficacy across different methods, and additionally, their respective trainable parameters. The results are shown in Table \ref{tab:mt-bench}.

\begin{table}[htb]
\centering
\scalebox{0.8}{
\begin{tabular}{c|l|c|c}
\hline
\textbf{Model} & \textbf{Method} & \textbf{Param(\%)} & \textbf{MT-Bench}\\
\hline
\multirow{8}{*}{\textbf{Llama-2-7b}}
& \textbf{LoRA} & {2.32} & {6.16}\\
& \textbf{DoRA} & {2.33} & {6.38}\\
& \textbf{rsLoRA} & {2.32} & {6.24}\\
& \textbf{AsyLoRA*} & {1.26} & {6.08}\\
& \textbf{AsyLoRA} & {2.47} & {6.20}\\
& \textbf{PiSSA} & {2.32} & {5.58}\\
& \textbf{MoRA} & {2.32} & {5.98}\\
& \textbf{BoRA} & {2.35} & \textbf{6.76}\\\hline
\multirow{3}{*}{\textbf{Mistral-7b-v0.1}} 
& \textbf{LoRA} & {2.26} & {6.75}\\
& \textbf{DoRA} & {2.28} & {6.68}\\
& \textbf{BoRA} & {2.30} & \textbf{6.81} \\\hline
\multirow{3}{*}{\textbf{Llama-3-8b}} 
& \textbf{LoRA} & {2.05} & {6.88}\\
& \textbf{DoRA} & {2.06} & {6.85}\\
& \textbf{BoRA} & {2.08} & \textbf{6.99}\\\hline
\end{tabular}
}
\caption{Average scores, on a scale of 1 to 10 (with higher scores indicating better performance), assigned by GPT-4 on the MT-Bench evaluation of answers generated by models Llama-2-7b, Mistral-7b-v0.1 and Llama-3-8b, fine-tuned using BoRA and several baseline methods. For the Llama-2-7b and Llama-3-8b models, we employed the hyperparameter settings specified in \citep{liu2024dora}. For Mistral-7b-v0.1 model, we optimized the hyperparameters for all methods. By AsyLoRA method we utilized a rank of 128 to achieve comparable numbers of trainable parameters, with * indicating the use of half the rank.}
\label{tab:mt-bench}
\end{table}

For the Llama-2-7b model, DoRA demonstrates a slight improvement in performance relative to LoRA, coupled with an marginally increase in training parameters. Additionally, DoRA surpasses other baseline methods. Notably, BoRA outperforms both LoRA and DoRA substantially, with only a slight augmentation in parameter count to 2.35\%.

In the cases of Mistral-7b-v0.1 and Llama-3-8b models, DoRA's performance is slightly inferior to that of LoRA. However, BoRA exhibits superior performance compared to both LoRA and DoRA.

Overall, BoRA consistently delivers superior performance across all examined models, establishing it as a more effective fine-tuning methodology compared to both LoRA and DoRA on this task.

\subsection{Performance on NLU tasks}

\begin{table*}[htb]
\centering
\scalebox{0.75}{
\begin{tabular}{c|l|c|ccccccccccc}
\hline
\textbf{Model} & \textbf{Method} & \textbf{Params(\%)} & \textbf{BoolQ} & \textbf{Winograde} & \textbf{Hellaswag} & \textbf{PIQA} & \textbf{SIQA} & \textbf{OBQA} & \textbf{ARC-e} & \textbf{ARC-c} &\textbf{avg.}\\
\hline
\multirow{3}{*}{\textbf{Llama-2-7b}}
& \textbf{LoRA} & {0.83} &\textbf{69.80} & {82.60} & {83.60} & {79.90} & \textbf{79.50} & {81.00} & {79.80} & {64.70} & {77.61}\\
& \textbf{DoRA} & {0.84} & {71.80} & {82.60} & {89.10} & {83.70} & {76.00} & \textbf{82.40} & \textbf{83.70} & {68.20} & {79.69}\\
 & \textbf{BoRA} & {0.85} & {69.79} & \textbf{83.27} & \textbf{93.16} & \textbf{83.90} & {78.05} & {79.40} & {83.21} & \textbf{69.71} & \textbf{80.06}\\\hline
\multirow{3}{*}{\textbf{Mistral-7b-v0.1}}
& \textbf{LoRA} & {1.16} & {73.49} & {85.87} & {95.11} & \textbf{88.19} & {80.04} & \textbf{88.00} & {88.26} & {74.74} & {84.21}\\
& \textbf{DoRA} & {1.16} & {74.28} & {86.42} & \textbf{95.54} & {88.03} & {79.89} & {85.00} & {87.92} & {75.68} & {84.10}\\
 & \textbf{BoRA} & {1.18} & \textbf{75.11} & \textbf{86.58} & {94.95} & {87.81} & \textbf{80.71} & \textbf{88.00} & \textbf{89.06} & \textbf{77.30} & \textbf{84.94}\\\hline
\multirow{3}{*}{\textbf{Llama-3-8b}}
& \textbf{LoRA} & {0.70} & {70.80} & {84.30} & {91.70} & {85.20} & {79.90} & {79.00} & {84.20} & {71.20} & {80.79}\\
& \textbf{DoRA} & {0.71} & {74.60} & {85.60} & {95.50} & {89.30} & {79.90} & {85.80} & {90.50} & {80.40} & {85.20}\\
 & \textbf{BoRA} & {0.72} & \textbf{76.54} & \textbf{89.42} & \textbf{96.66} & \textbf{90.10} & \textbf{82.24} & \textbf{88.20} & \textbf{93.27} & \textbf{83.28} & \textbf{87.46}\\\hline
\end{tabular}}
\caption{Comparison of accuracy metrics for Llama-2-7b, Mistral-7b-v0.1 and Llama-3-8b employing different PEFT techniques across eight commonsense reasoning datasets. The evaluation encompasses the entire test dataset, and the reported accuracies are in percentage form (*1\%). By llama-2-7b and Llama-3-8b model, we use the data reported in \citep{liu2024dora}. By Mistral-7b-v0.1, we select the best hyperparameters for all methods.}
\label{tab:NLU tasks}
\end{table*}

In this section, our method is benchmarked against the LoRA and DoRA methods. For each task, we automatically calculate the models' accuracy on the full test dataset. The results are summarized in Table \ref{tab:NLU tasks}.

For the models Llama-2-7b and Llama-3-8b, DoRA outperforms LoRA, while BoRA surpasses both LoRA and DoRA. For the model Mistral-7b-v0.1, DoRA's performance is slightly inferior; however, BoRA still ranks at the top.

In conclusion, although LoRA and DoRA serve as robust baselines, BoRA’s slightly higher parameter utilization leads to superior overall performance metrics. This indicates that, for these tasks, BoRA is likely the most advantageous PEFT method to employ.

\subsection{Ablation Study}

\subsubsection{Ablation of Training Parameters}

By rank 64, the trainable parameters of BoRA nearly match 101\% of those of LoRA. To eliminate the potential confounding factor of improved performance simply due to increased trainable parameters, we conducted an ablation study. In this study, we controlled the number of trainable parameters such that: LoRA(r=66) > DoRA(r=65) > BoRA(r=64).

As demonstrated in Table \ref{tab:trainable parameters}, despite LoRA having a higher number of trainable parameters compared to DoRA and BoRA, BoRA still exhibits superior performance.

\begin{table}[t]
\centering
\scalebox{0.8}{
\begin{tabular}{l|cc|c}
\hline
\textbf{Method} & \textbf{Rank} & \textbf{Params(\%)} & \textbf{Mt-Bench} \\
\hline
\textbf{LoRA} & {66}& {2.39}  & {6.34}\\
\textbf{DoRA} & {65} & {2.37} & {6.29}\\
\textbf{BoRA} & {64} & {2.35} & \textbf{6.76}\\\hline
\end{tabular}
}
\caption{Average scores, on a scale of 1 to 10 (with higher scores indicating better performance), assigned by GPT-4 on the MT-Bench evaluation of answers generated by models Llama-2-7b. The number of trainable parameters are controlled as: LoRA(r=66) > DoRA(r=65) > BoRA(r=64)}
\label{tab:trainable parameters}
\end{table}

\subsubsection{Ablation of Direction}

In \citep{liu2024dora}, a reason for applying the magnitude matrix to column dimension instead of row is not provided.  A potential concern is that the row dimension might offer greater efficiency, thereby enabling BoRA to achieve superior performance. To address this, we applied the magnitude matrix exclusively in the row dimension while keeping all other hyperparameters constant. The comparative results of this modification are presented in Table \ref{tab:rev}. A notable finding is that the performance across two different dimensions remains comparable. The primary factor contributing to the superior performance of BoRA might still be its inherent symmetry.

\begin{table}[h]
\centering
\scalebox{0.8}{
\begin{tabular}{l|cc}
\hline
\textbf{Method} & \textbf{Params(\%)} & \textbf{Mt-Bench} \\
\hline
\textbf{DoRA} & {2.33}  & \textbf{6.38}\\
\textbf{DoRA($row$)} & {2.34} & {6.33}\\\hline
\end{tabular}
}
\caption{Average scores, on a scale of 1 to 10 (with higher scores indicating better performance), assigned by GPT-4 on the MT-Bench evaluation of answers generated by models Llama-2-7b. In comparing DoRA and DoRA($row$), the primary distinction lies in the direction of the magnitude matrix, while all other hyperparameters remain consistent.}
\label{tab:rev}
\end{table}

\subsubsection{Ablation of the number of trainable parameters}

\begin{table}
\centering
\scalebox{0.8}{
\begin{tabular}{c|cc|cc}
\hline
\textbf{Rank} & \textbf{BoRA} & \textbf{Param(\%)} & \textbf{LoRA} & \textbf{Param(\%)}\\
\hline
\textbf{4} & \textbf{5.93} & {0.19} & {5.89} & {0.15}\\
\textbf{16} & \textbf{6.44}& {0.63}  & {6.24} & {0.59}\\
\textbf{64} & \textbf{6.76} & {2.35} & {6.16} & {2.32}\\
\textbf{128} & \textbf{6.19} & {4.56} & {5.93} & {4.53}\\
\hline
\textbf{avg.} & \textbf{6.33} & {-} & {6.05} & {-} \\
\hline
\end{tabular}
}
\caption{Average scores on a 1-10 scale (where higher scores denote superior performance), as evaluated by GPT-4 using the MT-Bench framework. The scores pertain to the answers produced by the Llama-2-7b model when fine-tuned with the LoRA and BoRA methods, assessed across various ranks.}
\label{tab:rank}
\end{table}

Additionally, we examine the effects that variations in the number of trainable parameters have on performance metrics. We utilize the ranks at \{4, 16, 64, 128\}, with corresponding alpha values set at \{8, 16, 128, 256\} to optimize performance outcomes. 

The data presented in Table \ref{tab:rank} demonstrate that BoRA exhibits superior average performance across various rankings, which determine quantities of trainable parameters.

\section{Conclusion}
In this paper, we revisited the current limitations of LoRA(Low-Rank Adaptation) and DoRA(Weight-Decomposed Low-Rank Adaptation) and proposes an innovative approach, BoRA, to address these challenges. By utilizing two independent magnitude matrices for rows and columns, our method innovatively employs bi-dimensional magnitude adjustments to achieve enhanced magnitude training while maintaining symmetry across both dimensions. Our comprehensive analysis, supported by a robust set of metrics and extensive experimental validation, demonstrates that BoRA significantly surpasses state-of-the-art PEFT techniques, including the foundational LoRA and DoRA methods. This advancement paves the way for more efficient and effective fine-tuning in various machine learning models.

\section*{Limitations}

Despite the promising results achieved by our proposed method, several limitations should be acknowledged:

\begin{itemize}
    \item Limited Linguistic Scope: Our method was primarily experimented on languages with limited morphological complexity, such as English. Consequently, the generalizability of our approach to languages with rich morphological structures (e.g., Finnish, Turkish, or Arabic) remains uncertain and requires further investigation.
    \item Performance on Smaller-Scale Models: The performance of our method on smaller-scale models, specifically those with fewer parameters such as BERT(base) or RoBERTa(base), has not been thoroughly evaluated. Our current evaluation focuses exclusively on RoBERTa(base), trained using the LoRA, DoRA, and BoRA methods, specifically assessed on the CoLA task. The results are documented in section \ref{limit}. Further experiments are needed to validate its effectiveness across a broader range of model scales.
    \item Training Time: The overall training process of our method requires more time compared to methods such as LoRA. The additional time cost is primarily due to the time complexity of the calculations involved, which is reported in more detail in the Appendix\ref{limit}. 
\end{itemize}

Addressing these limitations in future work could involve extending experiments to a more diverse set of languages, optimizing the method for smaller-scale models, and seeking computational efficiencies to reduce training times.

\section*{Ethics Statement}

This research strictly pertains to advancements and methodologies within the field of Natural Language Processing (NLP). Throughout the development and testing of our method, we have adhered to ethical guidelines to ensure our research poses no additional risks or unintended consequences beyond its primary scope. Our methodology does not involve any use of sensitive or private data, nor does it infringe upon the rights and privacy of individuals.

We made conscientious efforts to avoid any form of bias in our algorithm and to ensure that the system remains fair and equitable. Furthermore, the applications of our method have been carefully considered to prevent any potential misuse or societal harm. Where applicable, we have also complied with all relevant legal and ethical standards governing data collection, usage, and dissemination.

Our research is driven by the goal of contributing positively to the NLP community and providing tools that can be used responsibly for furthering advancements in the field. As such, we believe that our method does not present any potential risks beyond the scope of typical NLP research endeavors.

\bibliography{coling_latex}

\appendix

\section{Appendix \label{app}}

\subsection{Hyperparameters}

For NLG and NLU tasks, we varies several mean hyperparameters and remain other settings still. The still hyperparameters can be found in table.\ref{tab:other hyperparameters}. 

For NLG tasks, on base model Llama-2-7b and Llama-3-8b, we use the same hyperparameters described in \citep{liu2024dora} for LoRA and DoRA and selected the best hyperparameterfor BoRA. For base model Mistral-7b-v0.1, we remain the trainable parameters and epochs, and selected the best hyperparameters for all methods. Detail can be found in table.\ref{tab:NLG}.

For NLU tasks, on base model Llama-2-7b and Llama-3-8b, instead of retraining, we use the result reported in \citep{liu2024dora}, and for BoRA, use the same hyperparameters described in \citep{liu2024dora} for LoRA and DoRA and selected the best hyperparameter for BoRA. For base model Mistral-7b-v0.1, we remain the trainable parameters and epochs, and selected the best hyperparameters for all methods. Detail can be found in table.\ref{tab:NLU}.

In the ablation study of the number of trainable parameters, we selected the best learning rate for each rank setting, the other hyperparameters are remain the same as those for NLG tasks. Detail can be found in table.\ref{tab:rank hyperparameters}.

\begin{table*}[p]
\centering
\begin{tabular}{l|ccc}
\hline
\textbf{Hyperparameter} & \textbf{Llama-2-7b} \& \textbf{Mistral-7b-v0.1} \& \textbf{Llama-3-8b}\\
\hline
\textbf{Optimizer} & {AdamW} \\
\textbf{Dropout} & {0.0} \\
\textbf{Warmup ratio} & {0.1} \\
\textbf{LR Scheduler} & {Cosine} \\
\hline
\end{tabular}
\caption{Still Hyperparameter configurations for all baseline methods / models on NLG and NLU tasks}
\label{tab:other hyperparameters}
\end{table*}

\begin{table*}[p]
\centering
\begin{tabular}{c|l|ccc}
\hline
\textbf{PEFT Method} & \textbf{Hyperparameter} & \textbf{Llama-2-7b} & \textbf{Mistral-7b-v0.1} & \textbf{Llama-3-8b}\\
\hline
\multirow{5}{*}{\textbf{LoRA}} 
 & {Rank $r$ / alpha $\alpha$} & \multicolumn{3}{c}{64 / 128} \\
 & {Learning Rate} & \multicolumn{3}{c}{4E-4} \\
 & {Batch Size / Accumulation Steps} & \multicolumn{3}{c}{4 / 4} \\
 & {Epochs} & \multicolumn{3}{c}{1} \\
 & {LoRA Target Modules} & \multicolumn{3}{c}{Q, K, V, O, Up, Down, Gate}\\\hline
\multirow{5}{*}{\textbf{DoRA}} 
 & {Rank $r$ / alpha $\alpha$} & \multicolumn{3}{c}{64 / 128} \\
 & {Learning Rate} & \multicolumn{3}{c}{4E-4} \\
 & {Batch Size / Accumulation Steps} & \multicolumn{3}{c}{4 / 4} \\
 & {Epochs} & \multicolumn{3}{c}{1} \\
 & {LoRA Target Modules} & \multicolumn{3}{c}{Q, K, V, O, Up, Down, Gate}\\\hline
\multirow{5}{*}{\textbf{BoRA}} 
 & {Rank $r$ / alpha $\alpha$} & \multicolumn{3}{c}{64 / 128} \\
 & {Learning Rate} & {4E-4} & {3E-4} & {4E-4} \\
 & {Batch Size / Accumulation Steps} & \multicolumn{3}{c}{4 / 4} \\
 & {Epochs} & \multicolumn{3}{c}{1} \\
 & {LoRA Target Modules} & \multicolumn{3}{c}{Q, K, V, O, Up, Down, Gate}\\\hline
\end{tabular}
\caption{Main Hyperparameter configurations for all baseline methods / models on NLG tasks}
\label{tab:NLG}
\end{table*}

\begin{table*}[p]
\centering
\scalebox{0.8}{
\begin{tabular}{c|l|ccc}
\hline
\textbf{PEFT Method} & \textbf{Hyperparameter} & \textbf{Llama-2-7b} & \textbf{Mistral-7b-v0.1} & \textbf{Llama-3-8b}\\
\hline
\multirow{5}{*}{\textbf{LoRA}} 
 & {Rank $r$ / alpha $\alpha$} & {-} & {32 / 64} & {-}\\
 & {Learning Rate} & {-} & {3E-4} & {-} \\
 & {Batch Size / Accumulation Steps} & {-} & {16 / 1} & {-} \\
 & {Epochs} & {-} & {3} & {-} \\
 & {LoRA Target Modules} & {-} & {Q, K, V, O, Up, Down, Gate} & {-}\\\hline
\multirow{5}{*}{\textbf{DoRA}} 
 & {Rank $r$ / alpha $\alpha$} & {-} & {32 / 64} & {-}\\
 & {Learning Rate} & {-} & {2E-4} & {-} \\
 & {Batch Size / Accumulation Steps} & {-} & {16 / 1} & {-} \\
 & {Epochs} & {-} & {3} & {-} \\
 & {LoRA Target Modules} & {-} & {Q, K, V, O, Up, Down, Gate} & {-}\\\hline
\multirow{5}{*}{\textbf{BoRA}} 
 & {Rank $r$ / alpha $\alpha$} & {32 / 64} & {32 / 64} & {32 / 64} \\
 & {Learning Rate} & {2E-4} & {2E-4} & {1E-4} \\
 & {Batch Size / Accumulation Steps} & {16 / 1} & {16 / 1} & {16 / 1} \\
 & {Epochs} & {3} & {3} & {3} \\
 & {LoRA Target Modules} & {Q, K, V, Up, Down} & {Q, K, V, O, Up, Down, Gate} & {Q, K, V, Up, Down}\\\hline
\end{tabular}}
\caption{Main Hyperparameter configurations for all baseline methods / models on NLG tasks}
\label{tab:NLU}
\end{table*}

\begin{table*}[p]
\centering
\begin{tabular}{c|c}
\hline
\textbf{Rank} & \textbf{Learning Rate}\\
\hline
\textbf{4} & {5E-4} \\
\textbf{16} & {5E-4} \\
\textbf{64} & {4E-4} \\
\textbf{128} & {4E-4} \\
\hline
\end{tabular}
\caption{Learning rates for each Rank settings in ablation study: Ablation of the number of trainable parameters}
\label{tab:rank hyperparameters}
\end{table*}

\subsection{Limitations \label{limit}}

\begin{table*}[p]
    \centering
    \begin{tabular}{c|c}
    \hline
        \textbf{Method} & \textbf{CoLA} \\\hline
        LoRA & 60.82  \\
        DoRA & 61.83  \\
        BoRA & \textbf{61.93}  \\\hline
    \end{tabular}
    \caption{Performance of models trained on the CoLA dataset using three Parameter-Efficient Fine-Tuning (PEFT) methods: LoRA, DoRA, and BoRA. All models were trained with identical hyperparameters as specified in \citep{hu2021lora}. Each model underwent three independent runs, and we reported the average performance based on the best epoch from each run. For the CoLA dataset, we assessed model performance using the Matthews correlation coefficient in percentage form (*1\%).}
    \label{tab:cola}
\end{table*}

\begin{table*}[p]
    \centering
    \begin{tabular}{c|c}
    \hline
        \textbf{Method} & \textbf{Training Time} \\\hline
        LoRA & 139min \\
        DoRA & 176min \\
        BoRA & 179min \\\hline
    \end{tabular}
    \caption{Training duration per epoch for three PEFT methods on the Llama-2-7B model using the WizardLM-evol-instruct-70K dataset.}
    \label{tab:my_label}
\end{table*}

\end{document}